\pdfoutput=1

\documentclass[11pt]{article}

\usepackage{xcolor}
\usepackage{multirow}
\usepackage{float}

\usepackage{times}
\usepackage{latexsym}

\usepackage{hyperref}
\usepackage{url}
\usepackage{graphics}
\usepackage{graphicx}
\usepackage{caption}
\usepackage{subcaption}
\usepackage{amsmath}
\usepackage{amsfonts} 
\usepackage{amssymb}  
\usepackage{float}

\usepackage{bm}

\def\mC{{\bm{C}}}
\def\mD{{\bm{D}}}

\def\mX{{\bm{X}}}

\usepackage[final]{coling}

\usepackage[T1]{fontenc}

\usepackage[utf8]{inputenc}

\usepackage{microtype}

\usepackage{inconsolata}

\usepackage{graphicx}
\usepackage{amsmath}

\DeclareRobustCommand{\okina}{%
  \raisebox{\dimexpr\fontcharht\font`A-\height}{%
    \scalebox{0.8}{`}%
  }%
}

\title{Position Information Emerges in Causal Transformers Without Positional Encodings via Similarity of Nearby Embeddings}

\author{Chunsheng Zuo \\
  Dept. of Computer Science \\
  Johns Hopkins University \\
  \texttt{czuo3@jh.edu} \\\And
  Pavel Guerzhoy \\
  Dept. of Mathematics \\
  University of Hawai\okina i at M\=anoa \\
  \texttt{pavel@math.hawaii.edu} \\ \And
  Michael Guerzhoy \\
  Division of Engineering Science \\ 
  University of Toronto\\
  \texttt{guerzhoy@cs.toronto.edu}}

\begin{document}
\maketitle
\begin{abstract}
Transformers with causal attention can solve tasks that require positional information without using positional encodings. In this work, we propose and investigate a new hypothesis about how positional information can be stored without using explicit positional encoding. We observe that nearby embeddings are more similar to each other than faraway embeddings, allowing the transformer to potentially reconstruct the positions of tokens. We show that this pattern can occur in both the trained and the randomly initialized Transformer models with causal attention and no positional encodings over a common range of hyperparameters.  
\end{abstract}


\section{Introduction}
\label{Introduction}
Recent results by ~\citet{haviv2022nope},~\citet{kazemnejad2023nopelearnspositions}, and ~\citet{chi2023latent_posinfo_in_nope} suggest that positional encodings are not necessary when training decoder-only Transformer language models. These results motivate our investigation of how Transformers might represent positional information without positional encodings.

As shown in~\citep{tsai2019equivariance,zuo2024breaking}, the non-causal attention mechanism is equivariant to the permutation of the input tokens --- the prediction for input token $n+1$ is invariant to permutations of tokens $1, 2, ..., n-1$.
Therefore, without positional encodings, the causal attention mechanism is required for the Transformer to consider the order of the input tokens. ~\citet{chi2023latent_posinfo_in_nope}  hypothesize that causal attention allows positional information to be stored using the variance (taken across the indices of the embedding vector --- essentially the norm) of the embeddings, which generally decreases for tokens at later positions. 
They argue that the variance will tend to decrease because, when using causal attention, embedding $n$ is computed using embeddings $1, 2, ..., n-1$ in the previous layer, whereas embedding $n+k$ will be computed using $k$ more input embeddings, leading to variance shrinkage for embedding $n+k$.


We identify a different possible way of representing positional information that also arises from the fact that embeddings at earlier positions are computed using fewer embeddings from the previous layer compared to those at later positions. Specifically, we observe that embeddings at nearby indices will tend to be more similar to each other (in the sense of cosine similarity). This property could, in principle, enable the reconstruction of a token's position. 




The rest of the paper is organized as follows. We briefly review the literature on causal attention's connection to storing position information in Section~\ref{back:attention}. We then describe the pattern of nearby embeddings' being more similar to each other that we refer to as the \textit{adjacency pattern}, which we later link to the storing of position information in the network. We then present theoretical observations that explain how and why the adjacency pattern arises across many contexts in Section~\ref{appendix:origin}. We confirm through experiments on synthetic data that the pattern we report appears in a variety of configurations, both in trained and untrained architectures that use causal attention, in Section~\ref{mech}. We demonstrate a range of synthetic tasks where the position is important in Section~\ref{3:tasks}. In Sections~\ref{rand_init_exp},~\ref{exp_diff_model}, and~\ref{exp_diff_init}, we demonstrate that the pattern of large cosine similarity between nearby embeddings shows up in a variety of settings, for both trained \textit{and} untrained models. In Section~\ref{variance}, we point out that \citet{chi2023latent_posinfo_in_nope}'s theory of the position information's being stored in the embedding variance is insufficient to explain what we observe in our experiments. In Section~\ref{probing}, we explore the extent to which position information is stored in different Transformer layers, and to what extent that information can be thought of as being stored in the variance and in the \textit{adjacency pattern}. We discuss our results in Section~\ref{discussion} and discuss some limitations in Section~\ref{limitations}.




\begin{figure*}[h]
	\centering
       \includegraphics[width=\linewidth]{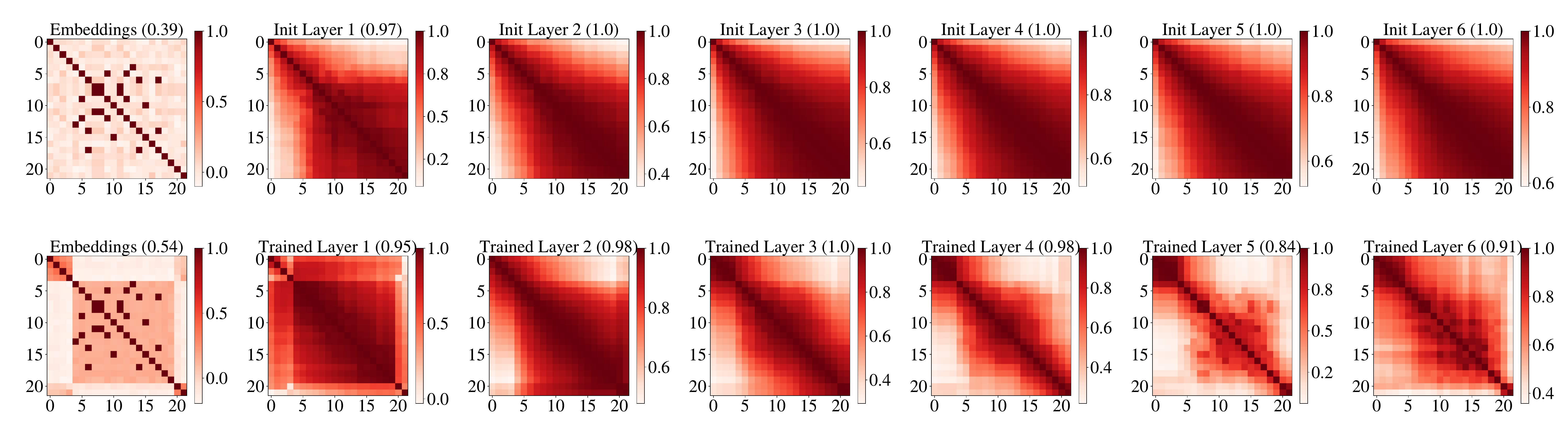}
		\caption{Self-cosine-similarity matrices of randomly initialized (first row) and trained (second row) 6-layer Transformers with causal attention and no positional encodings on the task of Reversal (22). The matrices are produced using a testing sample of 22 tokens, "rev(8502251258017069)=", as input, showing results from the embeddings to the output of layer 6 left to right for the initialized and trained models. The number in the bracket represents the adjacency probability score.}
	\label{fig:rev16}
\end{figure*}

\section{Background}


\subsection{Transformers with causal attention store position information without positional encodings \label{back:attention}}

Mechanisms analogous to modern attention in Transformers have long been used in recurrent neural networks~\cite{bahdanau2014neural, schmidhuber1992learning}. An attention mechanism is central to the Transformer architecture~\cite{vaswani2017attention}. 

In a Transformer with ``non-causal" attention, an output at the $k$-th position is agnostic to permutations in the positions of the inputs from other positions, a property known as permutation invariance (the logits above positions $1...k$ are permutation-\textit{equivariant} in the input). Without positional encodings, permutation equivariance prevents the output of each layer from taking into account the position of input tokens. In contrast, \citet{tsai2019equivariance} show that Transformers with causal attention are not permutation-equivariant to the input sequence. This implies the possibility of the success of~\citet{haviv2022nope} in training causal Transformers without positional encodings --- ``non-causal" attention could not accomplish that. 

 \subsection{The self-cosine-similarity matrix and the adjacency pattern}
The self-cosine-similarity matrix is a method to visualize the similarity (in the sense of a small angle) between all pairs of vectors within a sequence of embeddings. To create a self-cosine-similarity matrix $\mC$ for a sequence of $n$ token embeddings $\mX \in \mathbf{R^{n\times d}}$ of dimension $d$, we define each entry $\mD_{ij}$ as the cosine similarity between the $i^{th}$ and $j^{th}$ token embeddings, namely, $\mD_{ij}=\text{similarity}(\mX_i, \mX_j) = \cos \theta_{(\mX_i, \mX_j)}$ . Since the cosine similarity operation is commutative, $\mD_{ij}=\mD_{ji}$, resulting in the self-cosine-similarity matrix's being diagonally symmetrical. 

We use the term \textit{adjacency pattern} to describe a special type of self-cosine-similarity matrix that we observe. An example of this pattern can be found in Figure~\ref{fig:rev16}, where the matrix is darker (higher values) closer to the diagonal and brighter (lower values) further away, indicating that each embedding vector is more similar to vectors closer to it and less similar to vectors further away from it. 
The key idea in this paper is that embeddings exhibit an \textit{adjacency pattern}, meaning position information may, in principle, be partially recoverable from them, as embeddings corresponding to spatially nearby positions tend to be more similar.


The self-cosine-similarity matrix is used in~\citep{wang2020position} to visualize various positional encodings, some of which, such as the sinusoidal embeddings, demonstrate the \textit{adjacency pattern}. In our work, the self-cosine-similarity matrix is applied to the causal attention's output embeddings directly in order to examine their adjacency pattern.

\section{How the adjacency pattern arises \label{appendix:origin}
}


\citet{chi2023latent_posinfo_in_nope} demonstrate that, in the first hidden layer of the causal Transformer, the variance of the individual coordinate values within embeddings goes down with token index $k$. They infer that information about the position is related the the variance of the embedding. \citet{chi2023latent_posinfo_in_nope} explain the decrease in variance by observing that embedding $k$ is computed using a larger and larger context as $k$ grows.

In this section, we use the observation to show that we should expect for the adjacency pattern to arise in the first layer after the learned token embeddings.


\subsection{Empirical evidence}

Embeddings at positions \(k-1\), \(k\), and \(k+1\) are computed using the linear combinations of the value vector sets \(\{e_1, e_2, \ldots, e_{k-1}\}\), \(\{e_1, e_2, \ldots, e_{k-1}, e_k\}\), and \(\{e_1, e_2, \ldots, e_{k-1}, e_k, e_{k+1}\}\), respectively, where $e$ are the embeddings.

\begin{figure*}[h]
	\centering
       \includegraphics[width=\linewidth]{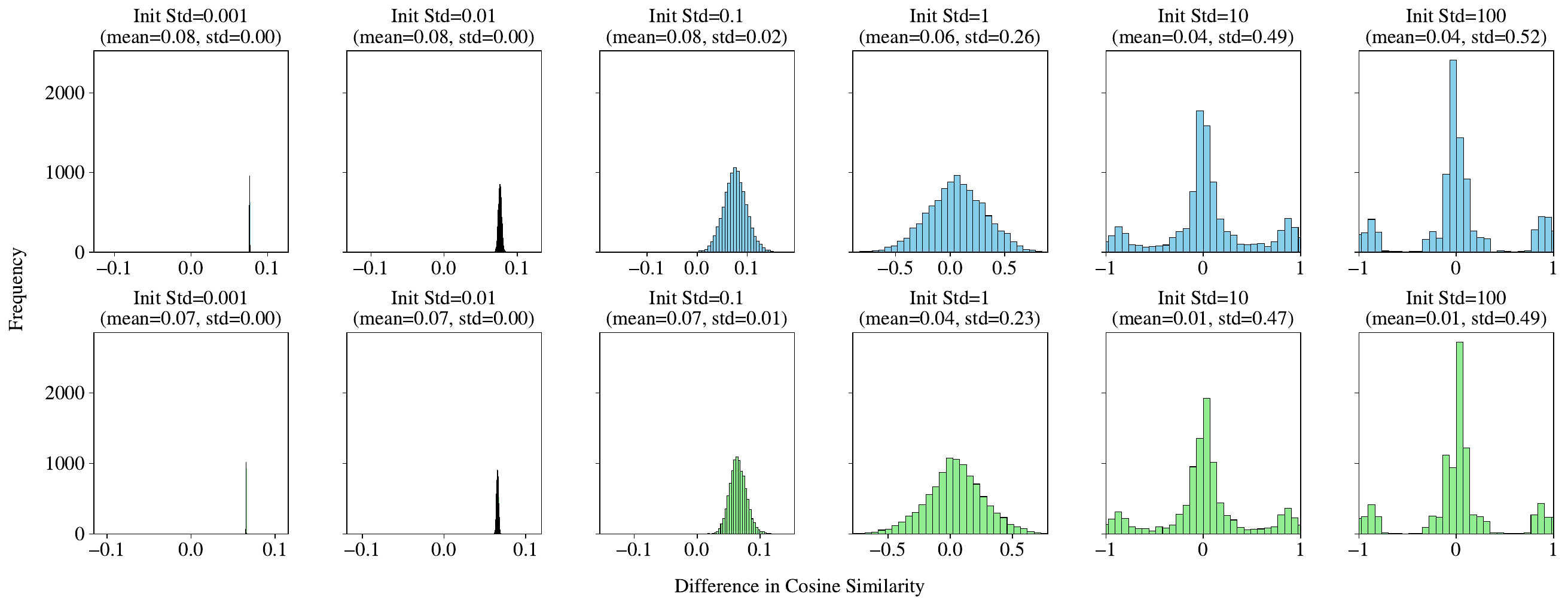}   
		\caption{Histograms on the differences between the cosine similarity of nearby tokens and further ones. Images in the first and the second row are for $sim(a, b) - sim(a, c)$, and $sim(c, b) - sim(c, a) $, respectively.}
	\label{fig:emperical_evidence}
\end{figure*}

We first simulate the value vectors used in attention by a set of random normal 128-dimensional vectors $\{v_1, ..., v_k\}$ and the causal attention weights at the 4th, 5th, and 6th row by the following i.i.d. random coefficient sets $\{\alpha_1, \alpha_2, \alpha_3, \alpha_4\}$, $\{\beta_1, \beta_2, \beta_3, \beta_4\}$, $\{\gamma_1, \gamma_2, \gamma_3, \gamma_4\, \gamma_5\, \gamma_{6}\}$. We then mimic the attention output embeddings at token positions 4, 5, and 6 by the following linear combination of vectors: $a=\left(\sum_{i=1}^{4} \alpha_i v_j, b=\sum_{i=1}^{5} \beta_i v_j, c=\sum_{i=1}^{6} \gamma_i v_j\right)$. Denote the cosine similarity as "$sim$". We want to determine the condition for $sim(a,b)$ to be consistently higher than $sim(a,c)$, as well as for  $sim(c,b)$ to be higher than $sim(c,a)$. We simulate with a range of standard deviations $\sigma_{init}$ from the set $\{0.001, 0.01, 0.1, 1, 10, 100\}$, and for each we repeat for 10000 trials and record $sim(a,b)-sim(a,c)$ and  $sim(c,b)-sim(c,a)$ for each trial. The resulting histogram is plotted in Figure~\ref{fig:emperical_evidence}, where the first and second rows are for $sim(a,b)-sim(a,c)$ and  $sim(c,b)-sim(c,a)$, respectively. The distribution is narrow and above zero for only small values of $\sigma_{init}$, corresponding to the condition that allows $sim(a,b)$ to be consistently higher than $sim(a,c)$ (same for $sim(c,b)$ and $sim(c,a)$). See also the experimental results in Table~\ref{tab:init_std}.


\subsection{The averaging effect provably arises in the first layer}

Here, we show that we can expect that, in the second layer (i.e., the first layer after the embeddings), the angle between embedding $k+t$ and embedding $k+t+1$ is smaller than the angle between embedding $k+t$ and embedding $k+t+2$, implying an adjacency pattern.

Assume embeddings $\{e_1, e_2, ..., e_k, ..., e_{n}\}$ are high-dimensional and normalized, and therefore approximately orthogonal. We are computing the next layer, with coefficients $\alpha$, $\alpha'$, $\beta$, and $\beta'$. 


We would like to show that the angle between $\sum_{i=1}^{k+t} \alpha_i e_i$ and $\sum_{i=1}^{k+t+1} \beta_i e_i$ would tend to be smaller than the angle between $\sum_{i=1}^{k+t} \alpha_i e_i$ and $\sum_{i=1}^{k+t+2} \beta'_i e_i$. The weights $\alpha$, $\beta$, and $\beta'$, which correspond to the attention weight in a causal architecture, would all sum to 1: $\sum_{i=1}^k \alpha_i = \sum_{i=1}^{k+t} \beta_i = \sum_{i=1}^{k+t+1} \beta'_i$ = 1.

Instead of the angles, we compute the dot products and show that we can expect the difference between the dot products to be positive, namely
\[
\left(\sum_{i=1}^{k+1} \alpha_i v_i \cdot \sum_{i=1}^{k+t} \beta_i v_i\right)  - 
\left(\sum_{i=1}^{k+1} \alpha_i v_i \cdot \sum_{i=1}^{k+t+1} \beta'_i v_i\right) > 0.
\]
Indeed, the difference between the left and right sides is

\begin{align*}
 &\;\;\;\;\; \sum_{i=1}^{k+1} \alpha_i v_i \cdot \sum_{j=1}^{k+t+1} (\beta_j - \beta’_j) v_j \\
 &\approx \sum_{i=1}^{k+1}  \alpha_i (\beta_i -\beta’_i) v_i \cdot v_i \\
 &\approx ||v|| \sum_{i=1}^{k+1} \alpha_i (\beta_i-\beta’_i) > 0,
\end{align*}

where the approximate equalities follow from the approximate orthogonality of large n-dimensional vectors normalized to norm 1.

\subsection{Semantic-level explanation}

If embedding $n$ is a summary of all information from positions $1..n-1$, we would expect that embeddings $n=k$ and $n=k+1$ be similar in \textit{some} space.

\section{The adjacency pattern appears in both non-trained and trained architectures in a variety of configurations \label{mech}}
In this section, we explore the settings in which the adjacency pattern in causal Transformers with no positional encodings (``Causal-NoPE") appears. We define the way we measure the adjacency pattern, describe the tasks we are using, and provide the experimental details.





\subsection{Adjacency probability score} \label{metric: adj pattern}

We propose the \textit{adjacency probability score} as a metric to quantify the ``intensity" of the adjacency patterns. The score is constructed to correlate with the amount of positional information that can be inferred from the self-similarity matrix.

We compute the proportion of time that the embeddings of tokens with closer positions have higher cosine similarity than those farther away, which can be derived directly from the self-cosine-similarity matrix. Consider the $k^{th}$ row of a squared matrix up to the column of the diagonal entry, denoted by $\mC_{k1}, \mC_{k2}, \dots, \mC_{kk}$. The row-wise adjacency probability score for this row is defined as:

\begin{align*}
P_{\text{Adjacency}} &= \mathbb{P}\left(\mC_{ki} < \mC_{kj} \text{ if } i < j \right)  \\
&= \frac{1}{{k \choose 2}} \sum_{j=0}^{i} \mathbb{I}\left(\mC_{ki} < \mC_{kj} \right)
\end{align*}

where $\mathbb{I}(\mC_{ki} < \mC_{kj})$ is 1 when $\mC_{ki} < \mC_{kj}$ and 0 otherwise. The adjacency probability score for the entire self-cosine-similarity matrix is calculated as the average row-wise adjacency probability score of all matrices. Notice that only the lower triangular portion of the matrix is involved in the calculation (see Appendix~\ref{appendix:metric}).

\subsection{Tasks} \label{3:tasks}

We trained Causal-NoPE Transformers for a variety of tasks that require positional information. The tasks were selected for being trainable from scratch and always requiring positional information.

\textbf{Addition:} \label{back:arithm} The Addition task involves generating the completion of strings like \texttt{"123+456="}. Following~\citet{lee2023teaching}, whose code base we also use, we train NanoGPT to generate the answer in reverse order. The input length (maximum and 90\% of the time) is 9 for 3-digit addition (we include strings like \texttt{"12+45"} as well).

\textbf{Reversal:} \label{back:reversal} The Reversal task requires the model to generate the reversed sequence. For example, for the prompt \texttt{"rev(1234)="}, the model is supposed to output \texttt{"4321"}. The input length (maximum and 90\% of the time) is 22  for reversing 16-or-less-digit numbers.

\textbf{Indexing:} \label{back:Indexing} The Indexing task requires the model to locate the position of the first occurrence of a number in the sequence. For an example, for the prompt \texttt{"wherex(134504392,4)="}, the model is supposed to output \texttt{"2"}, which is the index for the first occurrence of \texttt{"4"}. The input length (maximum and 90\% of the time) is 20 for indexing at most 9 digits.

\textbf{Ordering:} \label{back:ordering} Given a sequence of numbers and its reordered version, the Ordering task requires the model to output the new order of the original indices based on the reordered sequence. As an example, for the prompt \texttt{"order(67812,28716)="}, the model is supposed to generate the answer \texttt{"42130"}. The input length (maximum and 90\% of the time) is 18.
\subsection{Experimental Setup \label{experiments}}

We first want to examine whether the adjacency pattern persists for models trained for different tasks. We train the baseline 6-layer NanoGPT with 10.6 million parameters on each of the tasks. By default, all models are initialized by the normal distribution $\mathcal{N}(0,0.02)$. The training for each configuration is repeated for 5 different random seeds. Each task has 20000 training and 20000 testing samples. Otherwise, the configuration follows the work of \citet{lee2023teaching}, who trained the NanoGPT model to converge on the 3-digit Addition task. All experiments are conducted using an NVIDIA RTX4090 graphics card, with each trial being approximately 15 minutes.

Additionally, we want to compare the effect of different hyperparameters, particularly the number of layers and hidden dimensions. We choose the task of reversal and train models with 6, 12, and 24 layers and 192, 384, and 768 hidden dimensions, respectively, with the same train-test split. Unless further specified, the trained models have achieved more than 90\% accuracy in the testing set.

\section{Results}

\begin{table*}[h]
\centering
\renewcommand{\arraystretch}{1.3} 
\resizebox{\linewidth}{!}{%
\begin{tabular}{lccccccc}
\hline
Tasks & Embeddings & Layer 1 & Layer 2 & Layer 3 & Layer 4 & Layer 5 & Layer 6 \\
\hline
Addition (9) Init & $0.47$ & $0.99$ & $1.00$ & $1.00$ & $1.00$ & $1.00$ & $1.00$ \\
Addition (9) Trained & $0.48$ & $0.95$ & $0.98$ & $0.99$ & $0.98$ & $0.88$ & $0.85$ \\
Reversal (22) Init & $0.49$ & $0.97$ & $0.99$ & $0.99$ & $0.99$ & $0.99$ & $0.99$ \\
Reversal (22) Trained & $0.58$ & $0.91$ & $0.98$ & $0.99$ & $0.88$ & $0.82$ & $0.83$ \\
Indexing (20) Init & $0.49$ & $0.98$ & $0.99$ & $0.99$ & $0.99$ & $0.99$ & $0.99$ \\
Indexing (20) Trained & $0.55$ & $0.80$ & $0.96$ & $0.96$ & $0.88$ & $0.79$ & $0.83$ \\
Ordering (18) Init & $0.49$ & $0.98$ & $1.00$ & $1.00$ & $1.00$ & $1.00$ & $1.00$ \\
Ordering (18) Trained & $0.56$ & $0.89$ & $0.98$ & $0.96$ & $0.77$ & $0.80$ & $0.76$ \\
\hline
\end{tabular}
}
\caption{Averaged Layer-wise adjacency probability score for the 4 tasks, with initialization and trained results, each averaged over 256 samples. The number in the parentheses beside each task indicates the length (maximum and most frequent) of the equations in the task.}
\label{tab:performance}
\end{table*}

\begin{figure*}[h]
	\centering

       \includegraphics[width=\linewidth]{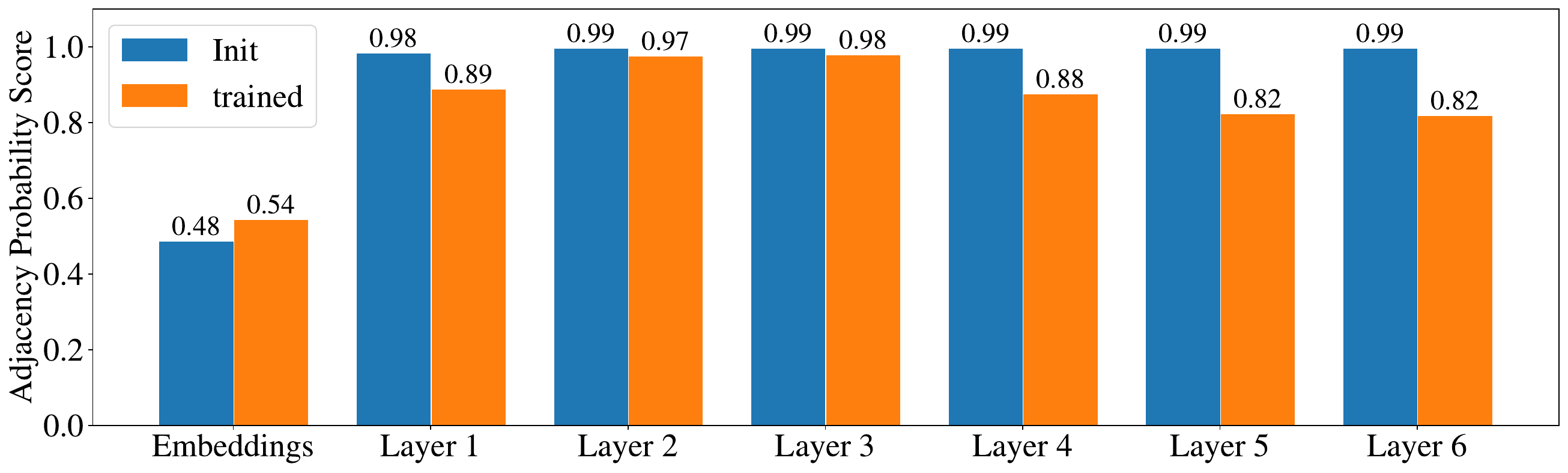}
		\caption{The layer-wise adjacency probability score for randomly initialized and trained models averaged over the 4 tasks, correspond to the values presented in Table~\ref{tab:performance}.}
	\label{fig:sum}
\end{figure*}

\begin{table*}[h]
\centering
\begin{tabular}{lccccccc}
\hline
Layers & Embeddings & Layer 1 & Layer 2 & Layer 3 & Layer n-2 & Layer n-1 & Layer n \\
\hline
6 &  $0.58$ & $0.91$ & $0.98$ & $0.99$ & $0.88$ & $0.82$ & $0.83$ \\
12 & $0.49$ & $0.90$ & $0.93$ & $0.96$ & $0.86$ & $0.81$ & $0.84$ \\
24 & $0.51$ & $0.84$ & $0.94$ & $0.84$ & $0.90$ & $0.78$ & $0.75$ \\
\hline
\end{tabular}
\caption{Layer-wise adjacency probability score for models with different numbers of layers trained on the Reversal (22) task, each averaged over 256 samples.}
\label{tab:layer_performance}
\end{table*}

\begin{table*}[h]
\centering
\begin{tabular}{lccccccc}
\hline
Dimensions & Embeddings & Layer 1 & Layer 2 & Layer 3 & Layer 4 & Layer 5 & Layer 6 \\
\hline
192 & $0.49$ & $0.93$ & $0.96$ & $0.96$ & $0.92$ & $0.81$ & $0.73$ \\
384 & $0.58$ & $0.91$ & $0.98$ & $0.99$ & $0.88$ & $0.82$ & $0.83$ \\
768 & $0.50$ & $0.96$ & $0.96$ & $0.95$ & $0.94$ & $0.90$ & $0.93$ \\
\hline
\end{tabular}
\caption{Layer-wise adjacency probability score for models with different numbers of hidden dimensions trained on the Reversal (22) task, averaged over 256 samples. The only configuration that did not achieve more than 90\% accuracy is the model with 192 dimensions, which has an accuracy of 56\%. Yet, we observed that in most cases where the model makes an error, the majority of digits are correct, with only a few being incorrect.}
\label{tab:dimension_performance}
\end{table*}

\begin{table*}[h]
\centering
\begin{tabular}{lccccccc}
\hline
$\mu_{init}$ & Embeddings & Layer 1 & Layer 2 & Layer 3 & Layer 4 & Layer 5 & Layer 6 \\
\hline
0 & $0.52$ & $0.97$ & $0.99$ & $0.99$ & $0.99$ & $0.99$ & $0.99$ \\
4 & $0.49$ & $0.97$ & $0.96$ & $0.95$ & $0.99$ & $0.96$ & $0.98$ \\
8 & $0.47$ & $0.97$ & $\textbf{0.56}$ & $\textbf{0.53}$ & $\textbf{0.57}$ & $\textbf{0.63}$ & $\textbf{0.65}$ \\
\hline
\end{tabular}
\caption{Layer-wise adjacency probability score for models initialized by Gaussian distribution with different means $\mu_{init}$, averaged over 256 samples from the Reversal (22) tasks.}
\label{tab:init_mean}
\end{table*}

\begin{table*}[h]
\centering
\begin{tabular}{lccccccc}
\hline
$\sigma_{init}$ & Embeddings & Layer 1 & Layer 2 & Layer 3 & Layer 4 & Layer 5 & Layer 6 \\
\hline
0.002 & $0.46$ & $0.97$ & $0.98$ & $0.98$ & $0.99$ & $0.98$ & $0.99$ \\
0.02 & $0.51$ & $0.97$ & $0.99$ & $0.99$ & $0.99$ & $0.99$ & $0.99$ \\
0.2 & $0.49$ & $\textbf{0.55}$ & $\textbf{0.58}$ & $\textbf{0.66}$ & $\textbf{0.70}$ & $\textbf{0.73}$ & $\textbf{0.68}$ \\
\hline
\end{tabular}
\caption{Layer-wise adjacency probability score for models initialized by Gaussian distribution with different standard deviation $\sigma_{init}$, averaged over 256 samples from the Reversal (22) tasks.}
\label{tab:init_std}
\end{table*}


\subsection{Transformer from random initialization knows positions right after the first causal attention \label{rand_init_exp}}

We computed the self-cosine-similarity matrix and the adjacency score across settings. Figure~\ref{fig:rev16} is representative of what we observe. In Figure~\ref{fig:rev16}, while there is no adjacency pattern in the matrices of the zeroth layer (i.e., the token embeddings), the adjacency pattern starts to appear in the output of the first attention layer and continues in the rest of the layers. The adjacency probability scores in the zeroth layer (i.e., the token embeddings) --- $0.39$ and $0.54$ for the randomly initialized and trained models respectively --- are much lower than in the other layers (where the minimum is 0.84). In those upper layers, the embeddings have been through at least 1 layer of causal attention. Hence, one layer of causal attention could be sufficient to generate the adjacency pattern.

\subsection{Adjacency pattern across different models and datasets \label{exp_diff_model} }

The adjacency probability scores of models trained for various tasks and with different hyperparameters (the number of hidden dimensions and the number of layers) are listed in Tables~\ref{tab:performance}, \ref{tab:layer_performance}, and \ref{tab:dimension_performance}. Each column of the table indicates the location where the embeddings are taken to produce the self-cosine-similarity matrices. Figure~\ref{fig:sum} presents the adjacency probability scores for the embeddings at each layer, averaged across different tasks. For Table~\ref{tab:layer_performance} and ~\ref{tab:dimension_performance}, the Reversal (22) task is chosen to demonstrate the effect of hyperparameters on the adjacency probability score. We observed that the effect of hyperparameters is the same across different tasks. 

For most configurations, the adjacency probabilities spike up from around 50\% in the token embeddings to more than 80\% at the first layer, as well as the rest of the layers. This is consistent regardless of the task type, the training state (initialized/trained), the number of layers, or the dimensions. As a general trend, the adjacency score is the highest for output embeddings in the second layer and declines gradually from there to the end. 

\subsection{The adjacency pattern across different initializations \label{exp_diff_init}}


We further test different initialization schemes, showing that the adjacency pattern is robust for the commonly used initialization schemes. Table~\ref{tab:init_mean} and Table~\ref{tab:init_std} show the results for the adjacency probability scores obtained in models initialized by Normal distribution with different means ($\mu_{init}\in\{0, 4, 8\}$) and different standard deviations ($\mu_{init}\in\{0.002, 0.02, 0.2\}$). The highlighted adjacency probability scores indicate a lack of discernible adjacency patterns qualitatively. The adjacency pattern is missing when the mean and the standard deviation are large enough ($\mu_{init}=4$ and $\sigma_{init}=0.2$), which are not typical values for initialization. It can be inferred that the mean has a smaller influence than the variance, since the first layer for the model with $\mu_{init}=4$ can still produce the adjacency pattern. Yet, it is likely that the large $\mu_{init}$ only causes the variance after the first layer to be large, which is why the adjacency pattern for the rest of the layers is removed.

\subsection{The variance alone may not be sufficient for accurate position information \label{variance}}

\begin{table*}[h]
\centering
\renewcommand{\arraystretch}{1.3} 
\resizebox{\linewidth}{!}{%
\begin{tabular}{lccccccc}
\hline
Tasks & Embeddings & Layer 1 & Layer 2 & Layer 3 & Layer 4 & Layer 5 & Layer 6 \\
\hline
Addition (9) Init & $0.45$ & $0.96$ & $0.98$ & $0.99$ & $0.99$ & $0.99$ & $0.98$ \\
Addition (9) Trained & $0.38$ & $0.93$ & $0.84$ & $0.92$ & $0.89$ & $0.84$ & $0.51$ \\
Reversal (22) Init & $0.47$ & $0.90$ & $0.95$ & $0.95$ & $0.97$ & $0.95$ & $0.97$ \\
Reversal (22) Trained & $0.66$ & $0.98$ & $0.61$ & $0.93$ & $0.88$ & $0.53$ & $0.63$ \\
Indexing (20) Init & $0.49$ & $0.94$ & $0.98$ & $0.99$ & $0.99$ & $0.99$ & $0.98$ \\
Indexing (20) Trained & $0.74$ & $0.93$ & $0.95$ & $0.98$ & $0.91$ & $0.90$ & $0.85$ \\
Ordering (18) Init & $0.55$ & $0.94$ & $0.99$ & $0.99$ & $0.99$ & $0.98$ & $0.98$ \\
Ordering (18) Trained & $0.22$ & $0.95$ & $0.76$ & $0.94$ & $0.71$ & $0.72$ & $0.43$ \\
\hline
\end{tabular}
}
\caption{Layer-wise adjacency probability score of the variance of embeddings for the 4 tasks, with initialization and trained results. The number in parentheses beside each task indicates the input length involved in the task.}
\label{tab:performance_of_norm}
\end{table*}

\begin{figure*}[h]
    \centering

    \begin{subfigure}[b]{\textwidth}
        \centering
        \includegraphics[width=\linewidth]{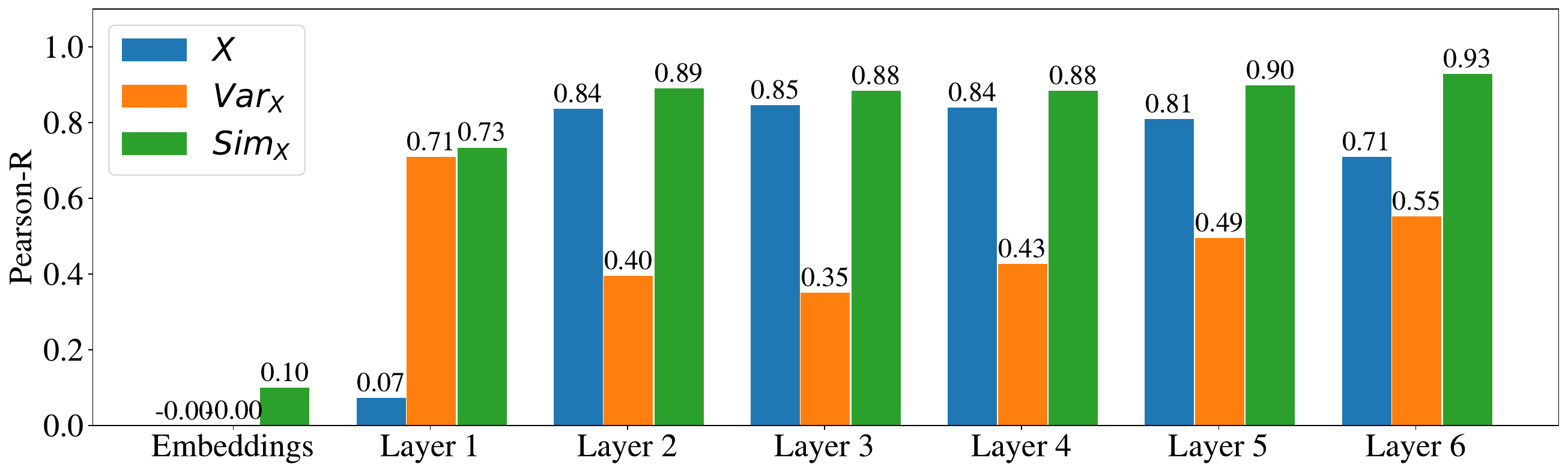}
        \caption{}
        
    \end{subfigure}
    
    \vskip\baselineskip
    \begin{subfigure}[b]{\textwidth}
        \centering
        \includegraphics[width=\linewidth]{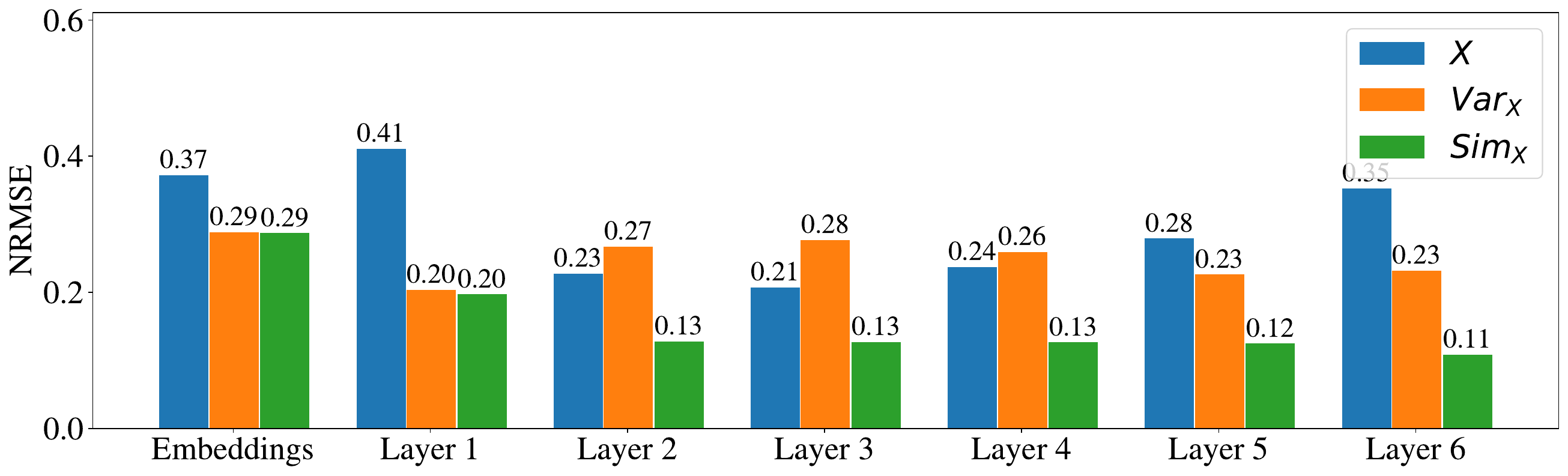}
        \caption{}
    \end{subfigure}
    \caption{Average layer-wise probing results for trained Causal-NoPE Transformers of (a) Pearson-R and (b) Normalized Root Mean Squared Error (NRMSE) using one of the following as the input: the output vector embeddings $X$, their variance $Var_X$, and the cosine similarity between the output vector embeddings and the vector at the last position $Sim_X$.}

    \label{fig:probing results}
    
\end{figure*}

\citet{chi2023latent_posinfo_in_nope} propose that the variance of the output embeddings tends to decrease from earlier to later positions, thereby serving as a signal of position information.
Hence, we applied the adjacency probability score to the variance of the embeddings to examine how well they are ordered, similar to what has been done to the self-cosine-similarity matrices. Given a sequence of embedding norms of length n, we repeat it n times to form a matrix and apply the same calculation in \ref{metric: adj pattern} to obtain the adjacency probability score. We also perform this evaluation for each task and put the results in Table~\ref{tab:performance_of_norm}. 

In the trained Causal-NoPE Transformers, there is a much more severe drop in the adjacency score of the norms than in the self-cosine-similarity matrices. Figure~\ref{fig:sum_norm} presents a visualization of the average results for Table~\ref{tab:performance_of_norm} across the 4 tasks. Compared to Figure~\ref{fig:sum}, it is clear that the average adjacency probability scores of the norms for the trained model are lower than for the self-cosine-similarity matrix. As an example, Figure~\ref{fig:rev_app} and Figure~\ref{fig:rev_app_norm} show the self-cosine-similarity matrices and norms for the initialized and trained model on the same task (Reversal (22)) with the same input. Though the norms tend to be monotonically decreasing at the initialized layers, they are not necessarily ordered in the trained layers, with the last layer even showing a reversed order. Even in comparison with the self-cosine-similarity matrices at the trained layers, except for the first layer, the norms are generally worse at indicating clear position information than the adjacency matrices.

\subsection{Probing for position information \label{probing}}

We further compare variance to cosine similarity by the effectiveness of using them as a feature to probe the position information. For each layer of the Causal-NoPE, the probe is trained to predict the position of an attention output vector using one of the following features as input: the output vector embeddings itself, its variance, and the cosine similarity between an output vector embeddings and the vector at the last position. The probe is a 4-layer Multi-Layer Perception (MLP) with 3 ReLU activation functions in between. To prevent the probe from memorizing the samples \citep{hewitt-liang-2019-designing}, the training and testing datasets for probing consist of random digits from 5 to 9 and 0 to 4, respectively, which are all contained in the 4 synthetic tasks~\ref{3:tasks}. We fix the input length to 32 and the training and testing sample size to 1600 each. The Root Mean Squared Error normalized by the input length (NRMSE) and the Pearson-R values are presented in Figure~\ref{fig:probing results}. We verify the validity of the results, showing that when using the output embeddings as features, the probe's testing performance in the setting of the untrained Causal-NoPE Transformers embeddings is the worst (in Appendix~\ref{appendix:probe} Figure~\ref{fig:probing results init}). 

In almost all layers of the initialized and trained models, using the cosine similarity values as the input feature produces the best probing outcomes with the lowest NRMSE and highest Pearson-R value. In addition to the adjacency probability score results, this probing result further demonstrates the robustness of inter-token cosine similarity as a positional indicator. In comparison, the variance seems less informative. In trained Causal-NoPE Transformers, from layers 2 to 6, the Correlation Coefficients of probes produced from the variance is worse than from the embeddings. This implies that if the Causal-NoPE Transformers learn to synthesize some absolute position information, it should rely on some characteristics of the embeddings more than just the variance.

\section{Discussion \label{discussion}}
\subsection{Is the adjacency pattern unique to causal attention?}
\begin{figure*}[h]
	\centering

       \includegraphics[width=\linewidth]{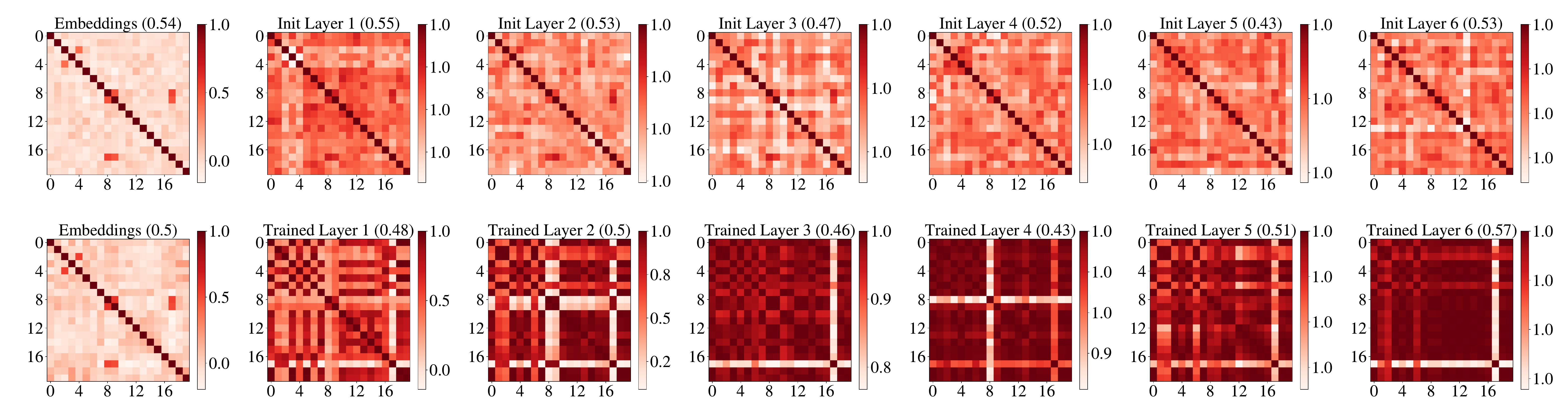}
		\caption{Self-cosine-similarity matrices of randomly initialized (first row) and trained (second row) 6-layer Transformers with normal attention and learned absolute positional encodings on the task of Indexing (20). The matrices are produced using a testing sample of 20 tokens, "wherex(299517340,9)=", as input.}
	\label{fig:wherex_nc}
\end{figure*}

Yes. We also applied a self-cosine-similarity matrix to Transformers with ``vanilla" attention and confirmed that there is no adjacency pattern. An example is shown in Figure~\ref{fig:wherex_nc}, where the self-cosine-similarity matrices look random and the adjacency scores are low. There is a learned absolute positional encoding added to the token embeddings of this model only to let the model converge.


\section{Limitations \label{limitations}}

The claims that the paper makes are partly based on empirical analyses of particular Transformer architectures, and using particular datasets. The observations would not necessarily generalize to other architectures. While an attempt was made to construct synthetic datasets that are interesting and display a variety of features, we do not mathematically prove that the observations we make would generalize to any dataset, and in fact it is likely that there could exist datasets to which our observations would not generalize.

\section{Conclusions and future work}
In Transformers with causal attention and no positional encodings, the adjacency pattern can occur for models with a wide range of hyperparameters, including the number of layers, hidden dimensions, and initialization schemes. It exists in the output embeddings of the Transformer's first causal attention layer and persists throughout the rest of the layers. For randomly initialized weights, the adjacency pattern can be observed for various initializations, especially for the ones commonly occurring in practice. For trained models, it is typical that the adjacency pattern in the first few layers is more prominent than in later ones, which we consider reasonable because knowing enough position information in the earlier layers may allow the models to focus on other more contextual information required by the tasks in later layers. 

Neither the adjacency pattern nor the in-embedding variance of~\cite{chi2023latent_posinfo_in_nope} can likely fully account for the fact that we are able to obtain 100\% performance on position-sensitive tasks since the probing results indicate that both are not 100\% informative. Nevertheless, we believe that the adjacency pattern provides another piece of the puzzle.

\section*{Acknowledgments}
We thank Prof. Ran Gilad-Bachrach for useful discussion. We thank Prof. Jonathan Rose for the conversation that initiated this investigation.

\bibliography{main}

\appendix

\section{More about the adjacency probability score}
\label{appendix:metric}
We only consider each row up to the diagonal because, for causal attention, each self-cosine-similarity matrix $S\in \mathbb{R}^{n\times n}$ of size n contains n sub-matrices, from $S_1\in \mathbb{R}^{1\times 1}$ to $S_n\in \mathbb{R}^{n\times n}$. For a sub-matrix of length $k\in[1,..,n]$, it is formed by embeddings resulting exactly from the first $k$ out of $n$ tokens of the original sequence. Therefore, each row-wise adjacency probability score at row $k$ measures the last row of sub-matrix $S_k$. Another way to think of this is that causal attention at the current token only considers anything before it. Hence, we measure just the adjacency probability score for anything up to the current token, which is up to the diagonal of each row.

\begin{figure*}[h]
	\centering
       \includegraphics[width=\linewidth]{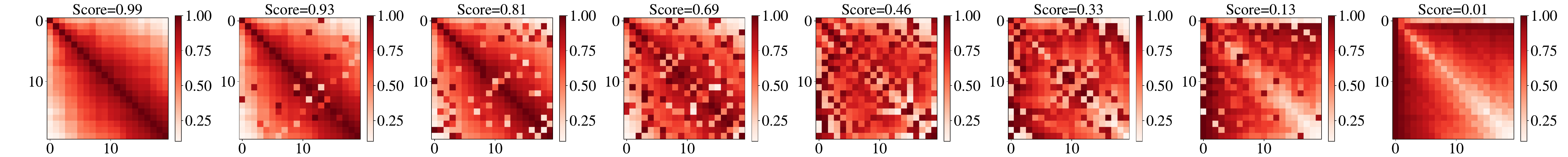}
		\caption{Synthetic matrices with different adjacency probability score values. (See Appendix~\ref{appendix:metric}}
	\label{sample:adj score}
\end{figure*}

Figure~\ref{sample:adj score} demonstrates different adjacency probability scores with their respective sample matrix. A higher adjacency probability score can be interpreted as the model being more likely to know the exact ordering of other tokens before a certain token. Meanwhile, although a zero adjacency probability score will also allow the model to know the token order oppositely, it is unachievable in a self-cosine-similarity matrix unless all embeddings are the same. For random matrices, the adjacency probability score is about 0.5.

\begin{figure*}[h]
	\centering
       \includegraphics[width=\linewidth]{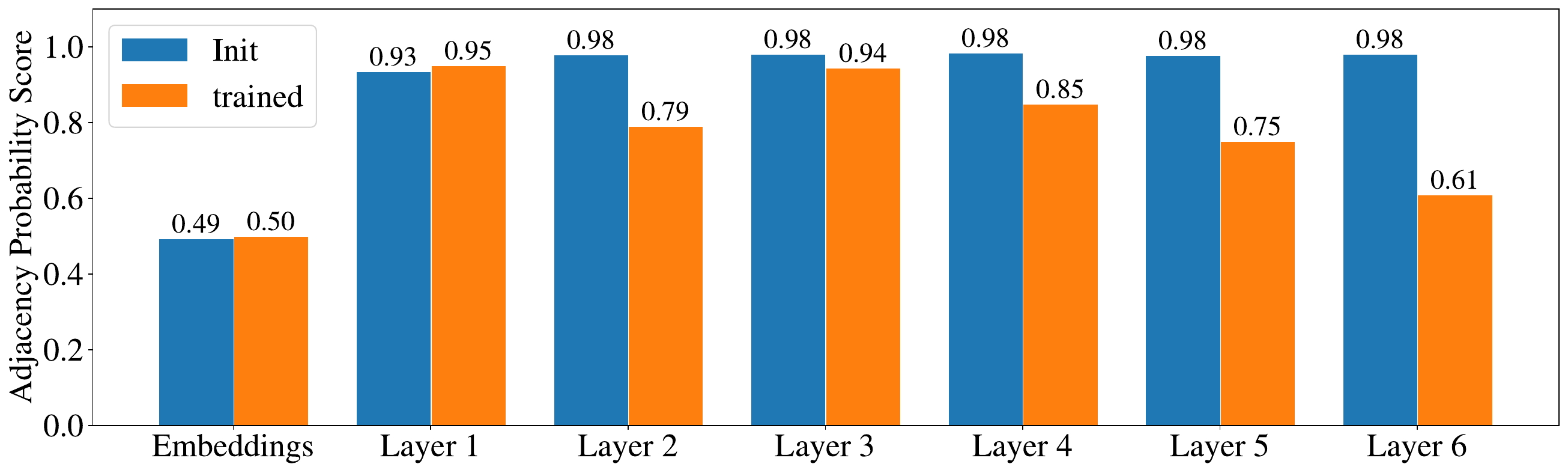}
		\caption{The layer-wise adjacency probability score of the norms for randomly initialized and trained models averaged over the 4 tasks, correspond to the values presented in Table~\ref{tab:performance_of_norm}.}
	\label{fig:sum_norm}
\end{figure*}

\section{Visualizations for probing results} 
\label{appendix:probe}
\begin{figure*}[h]
    \centering
    \begin{subfigure}[b]{\linewidth}
        \centering
        \includegraphics[width=\linewidth]{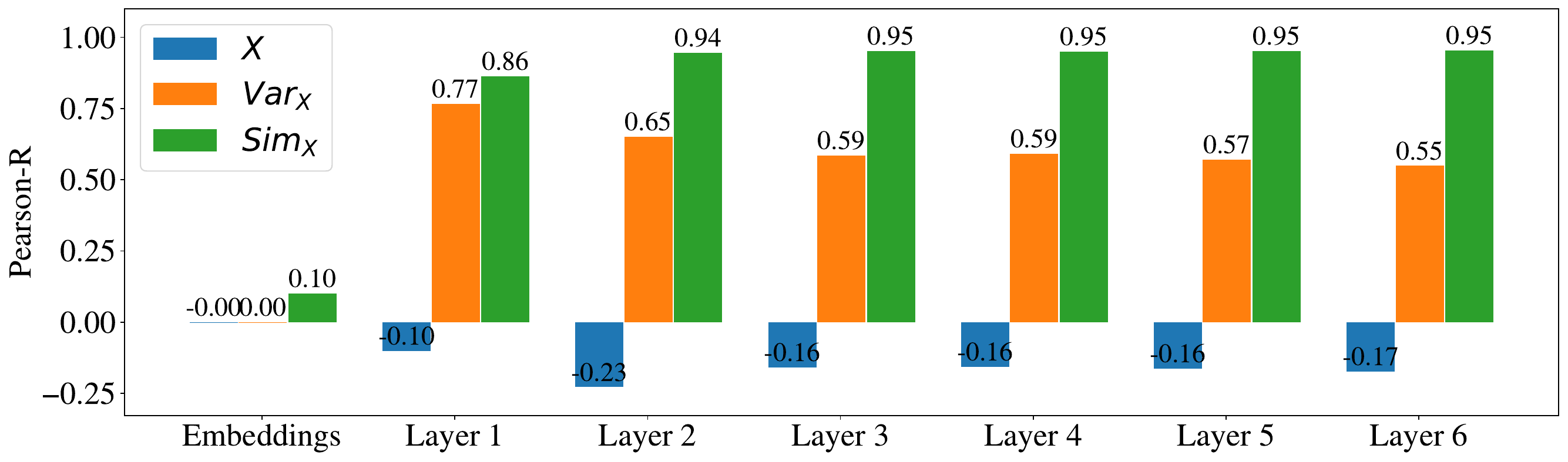}
        \caption{}
    \end{subfigure}
    
    \vskip\baselineskip
    \begin{subfigure}[b]{\linewidth}
        \centering
        \includegraphics[width=\linewidth]{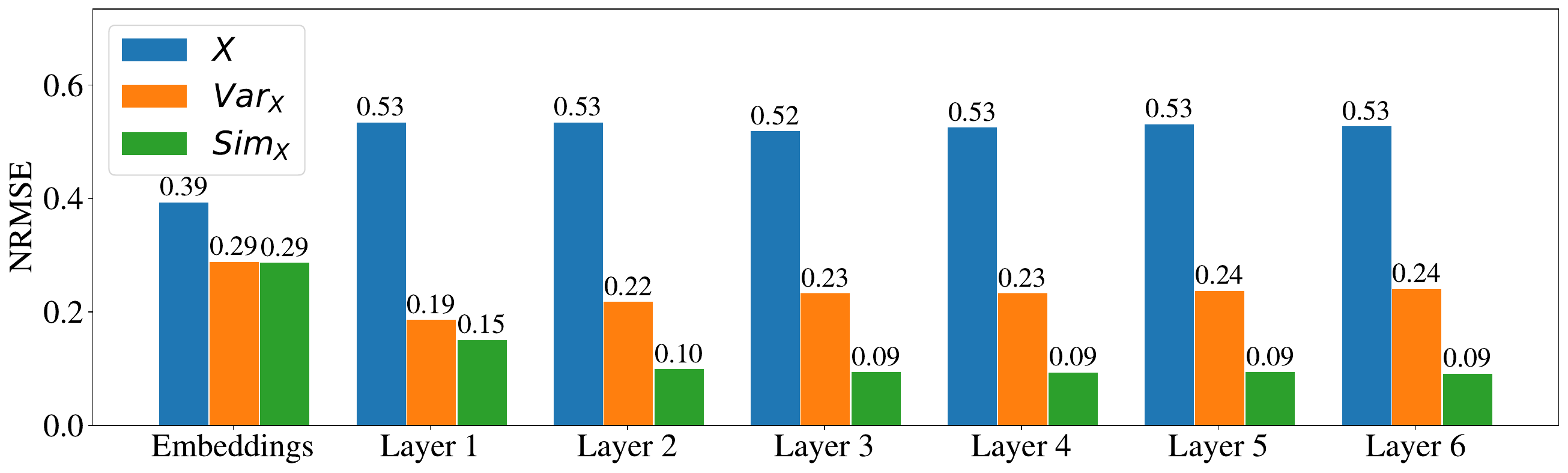}
        \caption{}
    \end{subfigure}
    \caption{Average layer-wise probing results for initialized Causal-NoPE Transformers of (a) Pearson-R and (b) Normalized Root Mean Squared Error (NRMSE) using one of the following as the input: the output vector embeddings $X$, their variance $Var_X$, and the cosine similarity between the output vector embeddings and the vector at the last position $Sim_X$.}

    \label{fig:probing results init}
    
\end{figure*}

Figure~\ref{fig:probing results init} shows the probing results on randomly initialized Causal-NoPE Transformers. The poor performance from using the embeddings as input indicates that the models do not contain any fixed/absolute positional information from the beginning, whereas the descent performance from using the cosine similarity as input suggests the existence of relative positional information inherent to the causal attention.

\begin{figure*}[h]
    \centering
    \vskip\baselineskip
    \begin{subfigure}[b]{\linewidth}
        \centering
        \includegraphics[width=\linewidth]{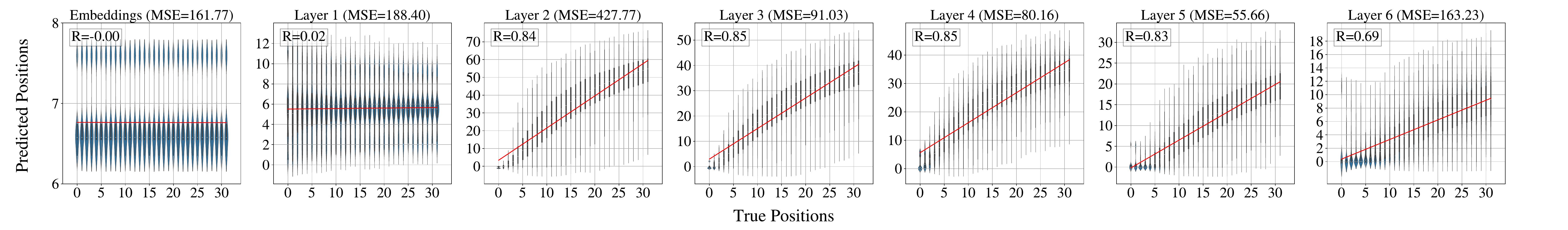}
        \caption{}
    \end{subfigure}

    \vskip\baselineskip
    \begin{subfigure}[b]{\linewidth}
        \centering
        \includegraphics[width=\linewidth]{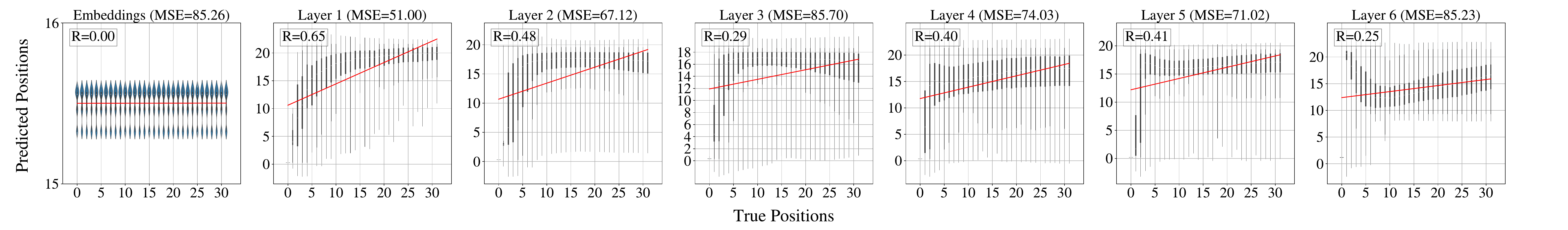}
        \caption{}
    \end{subfigure}

    \vskip\baselineskip
    \begin{subfigure}[b]{\linewidth}
        \centering
        \includegraphics[width=\linewidth]{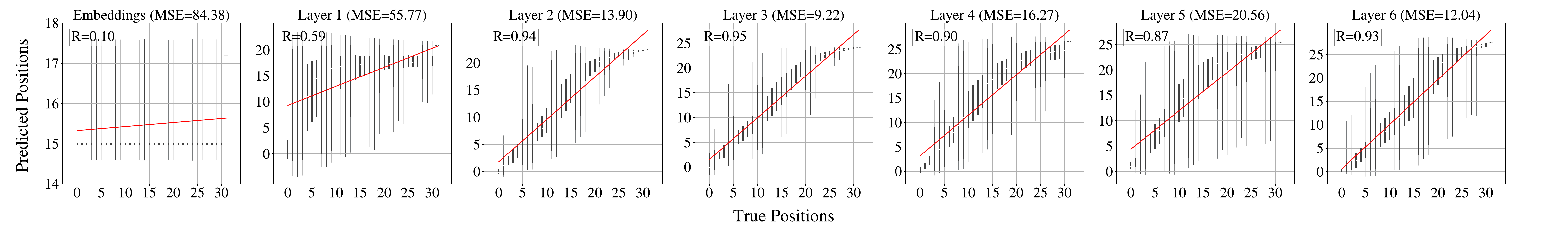}
        \caption{}
    \end{subfigure}
    \caption{Violin plots for the test predictions of a trained probe for a Causal-NoPE Transformer trained on the ordering task. The 3 different features, (a) embeddings, (b) variance, and (c) cosine similarity, are used independently.}
    \label{fig: probing details}
\end{figure*}

Figure~\ref{fig: probing details} demonstrates the prediction of probes trained using various input features of the trained Causal-NoPE Transformers.

\section{More Visualizations of Experimental Results}

\begin{figure*}[htbp!]
	\centering
       \includegraphics[width=\linewidth]{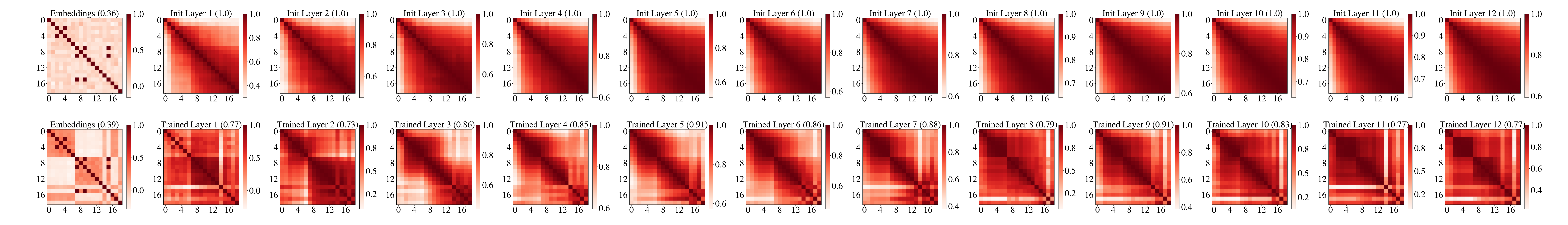}
		\caption{Self-cosine-similarity matrices of randomly initialized (first row) and trained (second row) 12-layer Transformers with causal attention and no positional encodings on the task of Indexing. The matrices are produced using a testing sample of 22 tokens, "wherex(8483561,8)=0", as input, showing results from the embeddings to the output of layer 12 left to right for the initialized and trained models. The number in the bracket represents the adjacency probability score.}
	\label{fig:rev16_24}
\end{figure*}

Figure~\ref{fig:rev16_24} provides an example of a model with 12 layers for the indexing task.

\begin{figure*}[h]
	\centering
       \includegraphics[width=\linewidth]{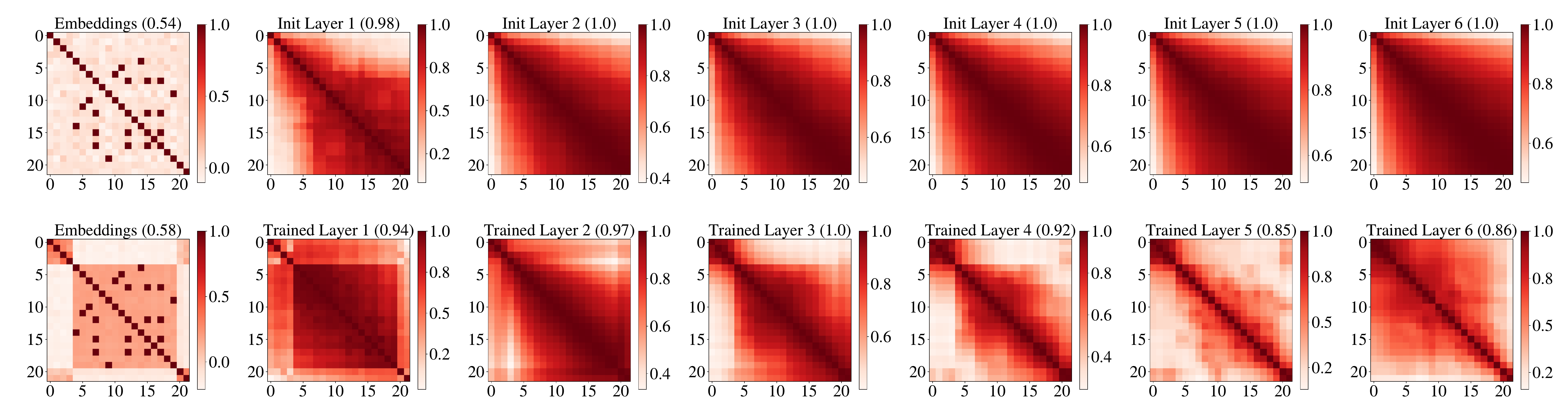}
		\caption{Layer-wise self-cosine-similarity matrices of randomly initialized (first row) and trained (second row) Causal-NoPE Transformers on the task of ordering, with "rev(1849364897192906)=" as the input.}
	\label{fig:rev_app}
\end{figure*}

\begin{figure*}[htbp!]
	\centering
       \includegraphics[width=\linewidth]{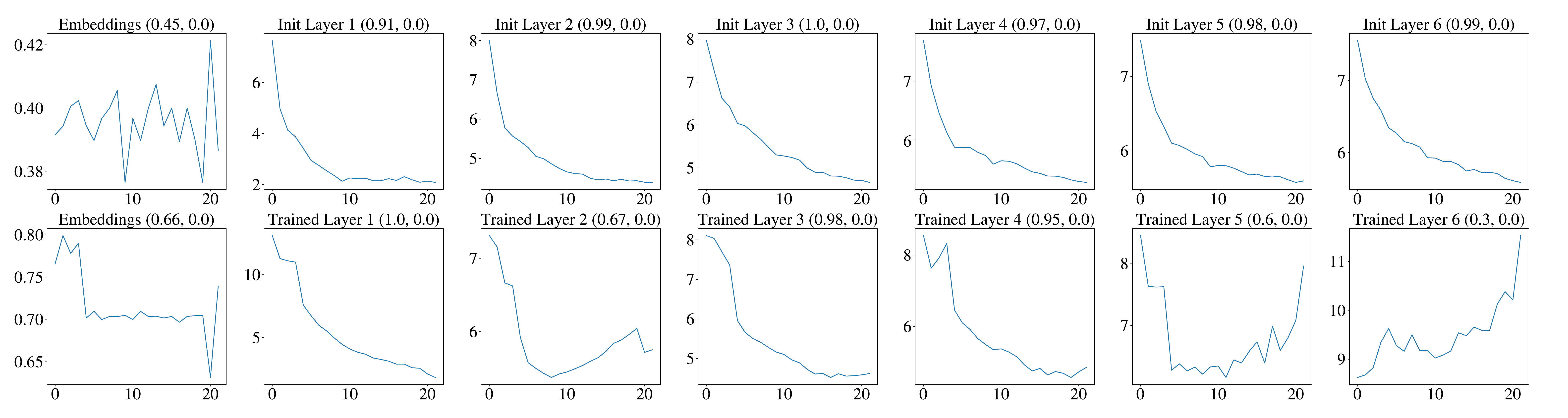}
		\caption{Layer-wise embedding norms for randomly initialized (first row) and trained (second row) Causal-NoPE Transformers on the task of Reversal (22), with "rev(1849364897192906)=" as the input.}
	\label{fig:rev_app_norm}
\end{figure*}

\begin{figure*}[h]
	\centering
       \includegraphics[width=\linewidth]{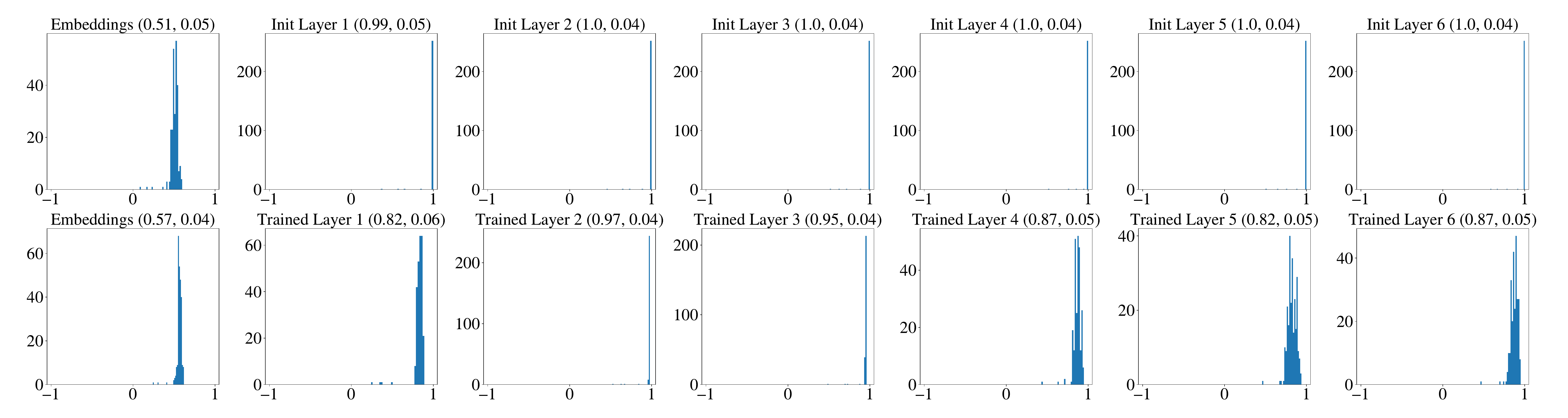}
		\caption{Distribution of adjacency probability score for a model before and after training ("Init"/"Trained") on the indexing task. The sample size of the histograms is 256. The two numbers inside the brackets of the subplot titles are the distribution's mean and standard deviation. Notice that the 7 pairs of means and standard deviations for the trained model (the second row) correspond to the values presented in Table~\ref{tab:performance} for the indexing task.}
	\label{dist:indexing}
\end{figure*}

 To determine if there are clusters of samples that exhibit extremely low to extremely high values, we check the distributions of the adjacency scores for all configurations. Typically, we observe distributions like the ones in Figure~\ref{dist:indexing} indexing task. In this example, while the distributions of adjacency scores concentrate around 1 for the untrained model, after training, only the adjacency scores for layer 2 and layer 3 distribute densely and closely to 1. In particular, the adjacency scores are the highest and most concentrated in layer 3 of the trained model, to an extent that matches the ones in the untrained model. We interpret these observations as an indication that the model learns to keep the adjacency pattern in earlier layers and gradually discard it in later ones. 

\end{document}